\def\year{2019}\relax
\documentclass[letterpaper]{article} 
\usepackage{aaai19}  
\usepackage{times}  
\usepackage{helvet}  
\usepackage{courier}  
\usepackage{url}  
\usepackage{graphicx}  
\usepackage{amsmath}
\usepackage{amsfonts}
\newcommand{\citet}[1]
{\citeauthor{#1} ̃\shortcite{#1}}
\newcommand{\citep}{\cite}

\begin{document}
%
\title{Text Assisted Insight Ranking Using Context-Aware Memory Network}
\author{
    Qi Zeng$^{1}$\thanks{Equal contribution.}, Liangchen Luo$^{2*}$, Wenhao Huang$^{3}$\thanks{This work was initiated and completed at the Software Analytics group of Microsoft Research Asia when the third author was full-time employee researcher and all the other authors were research interns of the group.}, Yang Tang$^2$\\
    $^1$Stony Brook University 
    $^2$Peking University 
    $^3$Shanghai Discovering Investment \\
    $^1${\tt qi.zeng@stonybrook.edu}
    $^2${\tt \{luolc,tangyang\_ty\}@pku.edu.cn}\\
    $^3${\tt huangwh@discoveringgroup.com}
}
\maketitle

\begin{abstract}
Extracting valuable facts or informative summaries from multi-dimensional tables, i.e. insight mining, is an important task in data analysis and business intelligence.
However, ranking the importance of insights remains a challenging and unexplored task. 
The main challenge is that explicitly scoring an insight or giving it a rank requires a thorough understanding of the tables and costs a lot of manual efforts, which leads to the lack of available training data for the insight ranking problem.
In this paper, we propose an insight ranking model that consists of two parts:
A neural ranking model explores the data characteristics, such as the header semantics and the data statistical features, 
and a memory network model introduces table structure and context information into the ranking process.
We also build a dataset with text assistance.
Experimental results show that our approach largely improves the ranking precision as reported in multi evaluation metrics.

\end{abstract}

\section{Introduction}

Automatically extracting useful and appealing insights, i.e. the data mining results, from a multi-dimensional table is a challenging yet important task in the areas of Business Intelligence (BI), Data Mining, Table-to-Text Generation, etc. 
For example, we can derive the insight "Sales of Brand A is increasing year over year while sales of Brand B is decreasing from 2015 to 2017 in China" from a multi-dimensional car sales table.
In this work, insight is defined as a data structure that includes subspace, type, significance value, and description. It can be described in any forms for different applications.
In the whole process of automatic business data analysis, generating abundant insights from multi-dimensional structured data can be accomplished with elaborate predefined rules, while modeling their usefulness or interestingness and ranking the top ones are much more difficult. 
Handcrafted ranking rules are less efficient and cannot cover every possible situation, and therefore a learning method for insight ranking is worth studying.

\begin{figure} 
\centering
\includegraphics[width=3.25in]{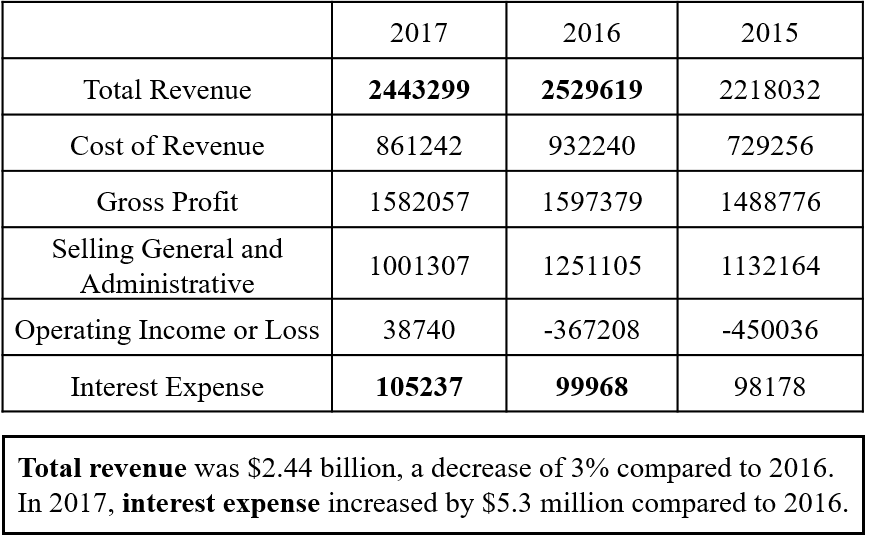}
\caption{Example of a table and its corresponding description text in an annual report.}
\label{fig:twitter}
\end{figure}

Previously effort has been made to explore how to extract insights according to its statistical significance score~\cite{tang2017extracting}. 
However, statistical significance has some limitation in insight importance ranking. 
First, as its scoring method suggests, it neglects the semantics of the data (such as the horizontal and vertical headers in the bi-dimensional table), which is proved to greatly contribute to the importance of data in our later experiments. 
As a result, insights that have a higher preference in real-world data are possible to get less attention. 
For example, in financial reports a statistically significant increase of ``Operating Income'' usually enjoys less popularity than that of a more common item ``Total Revenue'', as shown in Figure~\ref{fig:twitter}, but is possible to get a higher statistical significance score. 
Besides, the significance values of insights in different types are incomparable.
It is inaccurate to rank an insight of trend (shape insight) and an insight of outliers (point insight) according to their significant values since the two significant values have their own statistical meanings under different statistical hypothesis and measurements.
Also, the statistical analysis method is unsuited for small tables since it requires a minimum number of data points to calculate the statistical significance.

The main reason that the previous ranking methods are usually rule-based is that there is a lack of available training data for the insight ranking problem, as explicitly scoring an insight or giving it a rank usually requires domain knowledge and a thorough understanding of the table and context which is difficult and time-consuming. 
To address this problem, we take advantage of the human written table descriptions and analytical text, and use the text as "weak supervision" signals to learn an insight ranking model.
Such texts involve latent prior common knowledge and domain knowledge and provide valuable information on what insights are more important and are more likely to be mentioned.
To our best knowledge, this is the first work to explore the ranking problem of the insights with the assistance of text.

The importance of an insight can be measured in many dimensions. We find that the semantics information of insights contributes to its importance measuring.
The advantage of introducing it into the ranking model is that it provides the meaning to a cell of number, and the context of the data application.
Moreover, it breaks the limitation to table structure, as tables in any form and of any size can be universally represented as a list of labels and values. 
Inspired by this, in this paper we focus on ranking the insights by capturing both the semantic features and statistical characteristics of the data.
In addition,
the global table context, such as table structure and the relationship among all the insights, should also be taken into consideration.
For example, a year-over-year decreasing insight is more valuable than an increasing insight when all the other data are of increasing trends.
The challenges are three-fold. 
First, despite its prospect, there is no existing available dataset and no annotated insight importance labels for ranking models.
Second, it is hard to model the interestingness of insights as it can be measured in many dimensions. 
Both the content relevance and the statistical significance of insight need exploration.
Third, the comparison or ranking process among insights should be done in groups. 
For a fair comparison, insights within one table should be compared in one group since they are closely related inherently. 
Therefore, the table context needs to be introduced as external information to enable the comparison of relative interestingness values in a ranking model.


To overcome the above limitations, we present a text-assisted ranking model with header semantics and a global context-aware memory component.
We estimate the importance of an insight according to its probability of being interpreted in the description text and feed the score into the ranking model.
The ranking model consists of two parts.
The neural ranking model explores the data characteristics, such as its semantics and statistics information simultaneously. 
The key-value memory network model introduces table structure information into the ranking process.
The experiment results on two datasets demonstrate that our model achieves significant progress compared with baselines. 


In summary, our contributions are as follows:
\begin{itemize}
\item We formally formulate the problem of text assisted insight ranking, which has not been fully investigated yet.
\item We construct a new financial dataset, in which we labeled the insight importance with text assistance.
\item We propose a context-aware memory network to model the importance of insights.
The experimental results on two datasets show that our approach significantly outperforms the baseline methods.
\end{itemize}


\section{Related Work}

\subsection{Insight Ranking}

Earlier works have explored the insight importance evaluation problem. Notice that the insight has different names in different studies. A broader definition of the interestingness of insights, or data mining results, is conciseness, coverage, reliability, peculiarity, diversity, novelty, surprisingness, utility, and actionability \cite{geng2006interestingness}.

\citet{tang2017extracting} proposes that the insight score should be applicable to and fair across different types of insight. 
The insight score function in their paper measures the market share and the p-value based uncommonness significance score.
In their work, different insights follow different distribution and have different null hypothesis. 
We argue that such statistical methods do not satisfy the comparability requirement of insight importance score. 

\citet{demiralp2017foresight} also uses predefined strength metrics for each kind of insights, such as the Pearson correlation coefficient for linear relationship insight, the number of outliers for outliers insight, and standardized skewness coefficient for skew insight. 
More previous works in data exploration and data mining areas measure the insight importance by how surprising that value is different from the expectation \cite{wu2007towards,sarawagi1998discovery}. 
User preference is also taken into account in the area of interactive data exploration \cite{wasay2015queriosity,dimitriadou2016aide,cetintemel2013query}.
Different from their work, we introduce header semantics and table context into the insight ranking process.

In the task of table-to-text generation in Natural Language Processing (NLP), the generation process is divided into three modules, content planning, sentence planning, and surface realization~\cite{sha2017order,lebret2016neural,mei2015talk,liu2017table}. Similar to our insight ranking problem, the content planning module is required to decide which parts of the input table should be paid attention to. The difference is that the selection process is not explicitly formulated as a ranking problem that assigns each candidate a significance score.

\subsection{Learning to Rank}

The aforementioned insight importance ranking methods are mostly based on handcrafted rules, different from which our approach applies the ``learning to rank'' method in machine learning.

The ranking methods are usually classified into 3 categories, point-wise ranking, pair-wise ranking, and list-wise ranking.
The point-wise approach considers the ranking problem as multi-class classification problem \cite{DBLP:conf/nips/LiBW07} or regression problem \cite{DBLP:conf/colt/CossockZ06}. It considers the ranked candidates as independent, and is regardless of the final ranked result.
The pair-wise approach considers the ranking problem as binary classification problem and classifies the candidate pairs into two categories, correctly or incorrectly ranked pairs \cite{DBLP:conf/icml/BurgesSRLDHH05,DBLP:journals/jmlr/FreundISS03,DBLP:conf/nips/BurgesRL06,DBLP:conf/sigir/TsaiLQCM07}. 
There is a gap between its loss function and the evaluation metrics of the ranking results. 
The list-wise method scores the candidates within a list together and directly optimizes the evaluation metrics \cite{pareek2014representation,cao2007learning,DBLP:conf/nips/BurgesRL06,DBLP:conf/sigir/XuL07,DBLP:conf/wsdm/TaylorGRM08,burges2010ranknet}.

\section{Problem Formulation}




\subsection{Insight}



\textbf{\textsc{Definition 1 (Multi-Dimensional Table).}} A multi dimensional table is defined as the set of data cells, i.e. $T= \left\langle C_1, \cdots, C_c \right\rangle$. Each data cell $C_i$ is represented as $C_i = \left\langle Dim^1, \cdots, Dim^d, Val \right\rangle$, where $Dim^i$ is one dimension in a table, $d$ is the total number of dimensions in a table, and $Val$ is the value.

For example, table~\ref{tab:car} is a bi-dimensional table with dimension \textit{Brand} and \textit{Year}. For the cell in the up left corner,  $C_1= \left\langle Dim^1=\text{A}, Dim^2=2015, Val=13 \right\rangle$.

\begin{table}[!h]
\caption{Car Sales Table (Brand, Year, Sales)}
\centering
\begin{tabular}{c|ccc}
\hline
Brand, Year & 2015 & 2016 & 2017 \\
\hline
A           & 13   & 14   & 20   \\
B           & 51   & 49   & 60   \\
C           & 13   & 20   & 23   \\
\hline
\end{tabular}
\label{tab:car}
\end{table}

\textbf{\textsc{Definition 2 (Subspace).}}	A \textbf{subspace} is defined as a set of cells that
$S = \left\langle C_1, \cdots, C_n \right\rangle$, in which at least one dimension of the cells in the subset is the same:
\begin{equation}
\forall S = \left\langle C_1, \cdots, C_n \right\rangle,  \exists k  \text{ s.t. }  Dim^k_1 = \cdots = Dim^k_n,
\end{equation}
where $n$ is the number of cells in the subspace, and $Dim^k_i$ is the $k$-th dimension in each cell $C_i$.

In table~\ref{tab:car}, a subspace $S = \left\langle C_1, C_2, C_3\right\rangle$ consists of: 
\begin{equation}
 	\begin{split}
	 C_1&= \left\langle Dim^1=\text{A}, Dim^2=2015, Val=13 \right\rangle \\
	 C_2&= \left\langle Dim^1=\text{A}, Dim^2=2016, Val=14 \right\rangle \\
	 C_3&= \left\langle Dim^1=\text{A}, Dim^2=2017, Val=20 \right\rangle
     \end{split}
\end{equation}
where the cells share the same dimension $Dim^1=\text{A}$. The subspace is usually formed when we fixed some dimensions of the table and enumerate the combination of other chosen dimensions. The subspace usually has a particular meaning when selected. In the example, the subspace $S$ represents the sales of Brand A over years.

For each subspace, we can perform statistical test with specific hypothesis. 
The hypothesis is defined by \textbf{insight type $T$} which includes summary statistics, correlations, outliers, empirical distributions, density functions, clusters, and so on \cite{demiralp2017foresight}. 
Under the statistical hypothesis of insight type $T$, we can calculate the statistical \textbf{significance value $V$}. If the significance value exceeds a pre-defined threshold, it is considered as an informative observation from the table, and we can generate a \textbf{description $D$} from the header semantics for each dimension using some predefined templates, such as ``Sales of A is increasing from 2015 to 2017.'' in the example subspace we give in table~\ref{tab:car}.

Formally, we define the above elements as the insight:

\textbf{\textsc{Definition 3 (Insight).}}	An insight $I_i$ is defined as four parts
$I_i = (S_i, T_i, V_i, D_i)$, where $S_i$ is the subspace,  $T_i$ is the insight type, $V_i$ is the significance value, and $D_i$ is the corresponding description.

Table~\ref{tab:insightformat} is an example of an insight extracted from Table~\ref{tab:car}.

\begin{table}
\caption{Example of an insight}
\small
\centering
\begin{tabular}{l|l}
\hline
Subspace          	& $<$A,2015$>$, $<$A,2016$>$, $<$A,2017$>$   \\
\hline
Insight Type      	& Tread Increasing \\
\hline
Significance Value  & 0.5 \\
\hline
Description			& Sales of  A is increasing year over year. \\
\hline
\end{tabular}
\label{tab:insightformat}
\end{table}



\subsection{Text-Assisted Insight Ranking}

From a multi-dimensional table, we can derive a great many insights as informative observations especially when the table has many dimensions. However, people will only pay attention to several important insights, which requires insight ranking.
As introduced, it is difficult to explicitly calculate the importance of an insight directly. 
And human written table description and analysis text provide valuable information about what insights are more likely to be worth analyzing and which are not.

In this study, we use the assistance of the description text corresponding to a table.
Suppose the description is a set of sentences $\left\langle s_1, s_2, \cdots, s_m \right\rangle$.
For each insight $I_i$, we can calculate the similarity between the description $D_i$ from the insight and each sentence in the text by a similarity function $\mathrm{Sim}(D_i, s_j)$, and find the most similar sentence $s_k$.
When the similarity score is higher than a certain threshold, we can assume that human writer does mention that insight in the text, and correspondingly, the similar human written sentence is an expression of the insight. 
As a result, the semantic similarity score $\mathrm{Sim}(D_i, s_k)$ represents the possibility of an insight's being mentioned in the text, which further represents its importance or interestingness. 
The similarity $\mathrm{Sim}(D_i, s_k)$ can be seen as a "week supervision" of how likely the insight will be interpreted by people in the corresponding text.
Therefore, given an insight set $I=\left\langle I_1, I_2, \cdots, I_n \right\rangle$, we can get the rank of those insights by the similarity score as $\left\langle R_1, R_2, \cdots, R_n \right\rangle$.

\textbf{\textsc{Definition 4 (Insight Ranking).}} Given a set of insights $I$ from a table, we learn a \textbf{ranking function} $F$ 
\begin{equation}
 \begin{split}
&F:I_i \rightarrow \hat{R_i} \\
\text{s.t.}  &\min \sum_{i=1}^{n} L(R_i - \hat{R_i}) 
 \end{split}
\end{equation}
where $\hat{R_i}$ is the rank of insight $i$ from the ranking function, and $L$ is a list-wise loss function.

In order to compare all the insights from the same table together, we build our ranking function according to the list-wise ranking method in \cite{cao2007learning}.





\begin{figure*}[!htbp]
\centering
\includegraphics[width=5 in]{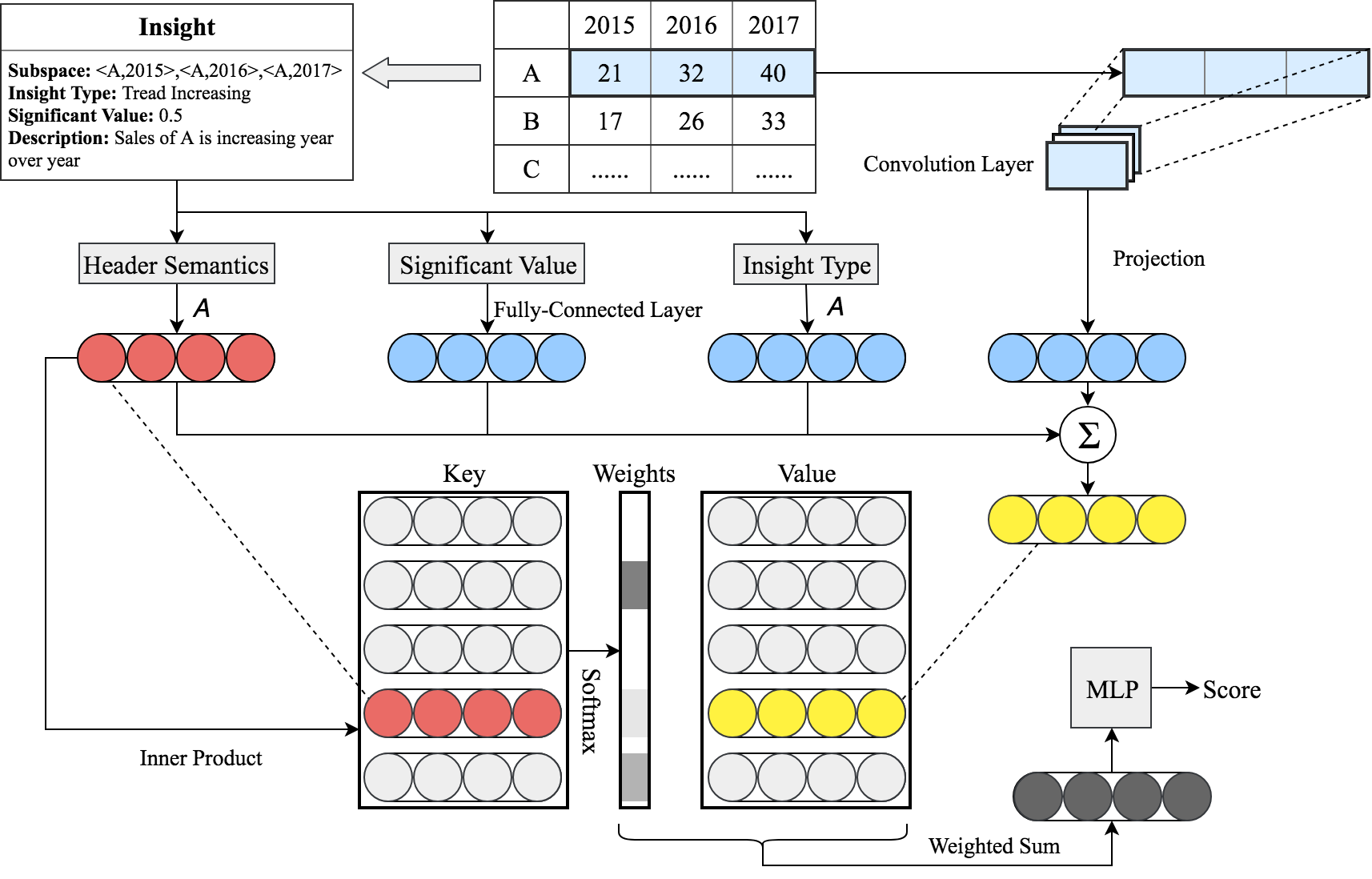}
\caption{Framework of the proposed ranking model.
}
\label{fig:model}
\end{figure*}

\section{Model}


As shown in Figure~\ref{fig:model}, our proposed model consists of two parts. The neural ranking model explores the data characteristics, including its semantic information, insight type, statistic information and subspace, and assigns importance scores to each insight. Additionally, the key-value memory network model introduces other insights within one group, namely the table context, into the ranking process.



\subsection{Insight Representation}
\label{model:insight}


We represent the insight with a vector of fixed size $d$.
Four kinds of insight features are encoded in different ways into vectors with the same vector length $d$.

\textbf{Significance value $f_{sig}$:} The significant score is embedded into a vector with a fully-connected layer.

\textbf{Insight Type $f_{type}$:} We treat each insight type as a special word token
and encode them using the same embedding matrix $A$ used in header semantics representation.

\textbf{Subspace $f_{subspace}$:}
The cells in a subspace is considered as a sequence of continuous cell values along with their shared dimension.
The sequence $C$ is then processed by a single-layer CNN to form the subspace representation.
The CNN regards $C$ as an input channel, and alternates convolution operation.

Suppose that $z^{(f)}$ denotes the output of feature maps of channel-$f$.
On the convolution layer, we employ a 1D convolution operation with a window size $r$, and define $z^{(f)}$ as:
\begin{equation}
	z^{(f)} = \sigma(\sum_{t=0}^r W_t^{(f)} \cdot C_t + b^{(f)}),
\end{equation}
where $\sigma(\cdot)$ is a tanh, 
$W^{(f)} \in 
\Re^r$
and $b^{(f)}$ are parameters.
The output of the feature maps are projected to a vector $f_{subspace}$ of dimension $d$ with a linear transformation.


\textbf{Semantics $f_{semantics}$:} The semantics of an insight is expressed as the concatenation of all headers of the cells in a insight subspace, which produces a sequence of word tokens
\begin{equation}
	x = [w_1, \cdots, w_h],
\end{equation}
where $h$ is the length of the headers.
Then the distributed semantics representation $s$ is defined as a bag-of-words using the embedding matrix $A$:
\begin{equation}
	f_{semantics} = A\Phi(x),
\end{equation}
where $\Phi(\cdot)$ maps the tokens to a bag of dimension $V$ (the vocabulary size), and $A$ is a $d \times V$ matrix.

Finally, the feature of an insight is represented by summing up the four features:
\begin{equation}
	I = f_{sig} + f_{type} + f_{subspace} + f_{semantics}.
\end{equation}



\subsection{Key-Value Memory Network}
\label{model:memory}



Our intuition is to introduce the table context such as table structure and the relations between insights in the same table into the ranking model.
We represent the table as a set of insights extracted from the table.


Since the insights are not naturally expressed as sorted sequence, a memory-like framework is more appropriate than structure-sensitive models such as RNN and CNN.
Assuming that relation between insights can be revealed by their header semantics, we apply a key-value memory network (KV-MemNN) \cite{DBLP:conf/emnlp/MillerFDKBW16} to search semantically similar insights for each insight candidate.

We define the memory slots as a vector of pairs
\begin{equation}
	m = [(s_1, I_1), \cdots, (s_M, I_M)],
\end{equation}
where there are $M$ related insights, $I_k$ is the $k$-th insight and $s_k$ is the semantic vector of insight $I_k$.
We denote the semantic of current insight as query $q$.
The key addressing and reading of the memory involves the following two steps.

\textbf{Key Addressing:}
During addressing, we perform a self-attention operation by computing the inner product between $q$ and the memory keys followed by a softmax:
\begin{equation}
	\alpha_i = \mathrm{Softmax}(q^\top s_i),
\end{equation}
which yields a vector of attention weights over the semantics of related insights.

\textbf{Value Reading:}
In the reading step, the values of the memories (insight representations) are read by taking their weighted sum using the addressing attentions, and the memory output vector $o$ is returned as:
\begin{equation}
	o = \sum_k \alpha_k I_k,
\end{equation}
The final insight representation $o$ will be an input of the ranking model described in the next section.





Since the representation of the insight itself is also contained in the memory, it will definitely produce very high attention to address the insight self.
We do not concatenate the output of the memory with other feature vectors as the other memory network often does.

\subsection{Ranking Model}
\label{model:loss}



The model is implemented as a multi-layer perceptron (MLP) which receives insight representations and outputs the ranking scores of the insights.

The model is trained by minimizing the L2 loss $J(\gamma)$ of the output scores and the similarity scores of the insights:
\begin{equation}
\begin{aligned}
score_m &= \mathrm{MLP}(o),\\
J(\gamma) &= \frac{1}{2}\|score_m - score_s\|_2^2,
\end{aligned}
\end{equation}
where $score_m$ and $score_s$ are the model outputs and ground-truth scores, respectively. We apply the list-wise approach and sums up the total losses of the insights in the same table, as the total loss relies on the table context. For the baseline models without the memory network, we apply the point-wise approach and calculate the L2 loss for each insight as a training sample.

\section{Dataset}


\subsection{Financial Report Dataset}
The financial report dataset
is built upon the public annual and quarterly reports from United States Securities and Exchange Commission\footnote{https://www.sec.gov/edgar/searchedgar/companysearch.html}. 
The dataset contains in total 5,670 reports and 49,129 tables from 2,762 companies. 
Table~\ref{tab:dataset_statistics} summarizes the data statistics. 
In the experiment, we randomly split the dataset into training, validation, and test sets consisting of 60\%, 20\%, and 20\% summaries, respectively.

We filtered the sentences out that are less than 50 characters or 10 words, and those do not include any numbers or keywords. 
Year information is substituted with ``this year'', ``last year'', and so on. 
More detailed date information is deleted as we only consider annual report. Special tokens are also processed to avoid noise.

\begin{table}[h]
\caption{Financial Report Dataset statistics.}
\centering
\begin{tabular}{l|c|cc}
\hline
  & \textbf{Mean} & \multicolumn{2}{c}{\textbf{Percentile}} \\
  &               & 5\% & 95\% \\
\hline
\# tokens per cell    &   5.29  &   1  &  12 \\
\# tokens per sentence      &  32.36   &  15   &  64  \\
\# sentences per report     &  774.98 &   282  &  1434 \\
\hline
\end{tabular}
\label{tab:dataset_statistics}
\end{table}

\subsection{SBNation Dataset}

To validate the generality of our model, we also evaluate
the effectiveness of our model 
on SBNation Dataset from \cite{DBLP:conf/emnlp/WisemanSR17}. 
This dataset consists of 10,903 human-written NBA basketball game summaries aligned with their corresponding box-scores and line-scores. 
We randomly split the dataset into training, validation, and test sets consisting of 60\%, 20\%, and 20\% summaries.

\subsection{Insight Extraction}

We defined two representative types of insight in this work:
\begin{itemize}
	\item{\textbf{Point insight:} we measure how outstanding the data point is among all the data points in the subspace.}
    \item{\textbf{Shape insight (trend):} we detect the rising or falling trend among a series of data points.}
\end{itemize}

In the financial dataset, the ``point'' is defined as the change ratio of one item from the current year to last year in the point insight.
The ``trend'' is defined as the increasing or decreasing trend year-over-year in the shape insight.

In the SBNation dataset, we only extract the point insight. The ``point'' is defined as the one of the game statistic such as scores, rebounds and assistants of a player.

The significance score of each insight type is calculated with the same approach described in \cite{tang2017extracting}.



\subsection{Similarity Function}
\label{similarity function}
We propose two similarity functions here, $\mathrm{Sim}_{s}$ and $\mathrm{Sim}_{s+h}$, and select the better one by human evaluation.




First, we count same words in the insight description $d_i$ and the human written sentence $s_j$:
\begin{equation}
\mathrm{Sim}_{s}(d_i, s_j) = \frac{\mathrm{Count}^2(d_i, s_j)}{|d_i| \cdot |s_j|}
\end{equation}
where $\mathrm{Count}(d_i, s_j)$ is the count of same words,
$|*|$ represents the length of $*$.

To assign more weights to the words in the headers, we calculate the similarity of a header $h_i$ and a sentence $s_j$:
\begin{equation}
\mathrm{Sim}_{h}(h_i, s_j) = \frac{\mathrm{Count}(h_i, s_j)}{|h_i|} \cdot \frac{\mathrm{Count}(h_i, s_j)}{\max_{k=1}^{n}\{\mathrm{Count}(h_k, s_j)\}} 
\end{equation}
where
$\frac{\mathrm{Count}(h_i, s_j)}{|h_i|}$ represents the percentage of the number of words matched in the header, and $\frac{\mathrm{Count}(h_i, s_j)}{\max_{i=1}^{n}\{\mathrm{Count}(h_k, s_j)\}}$ is the normalization factor.

We add the similarity of headers to the similarity of sentences to 
construct the second similarity function:
\begin{equation}
\mathrm{Sim}_{s+h}(\cdot) = \alpha_1 \mathrm{Sim}_{sent}(\cdot) + \alpha_2 \mathrm{Sim}_{h}(\cdot)
\end{equation}
where $\alpha_1$ and $\alpha_2$ are the weights. 
In this paper, we set them both to $0.5$.

Finally, the maximum similarity score among the insight description and all the candidate sentences in the text represents the probability of the insight's being interpreted in the report, which is further used as the guideline for ranking.

\subsection{Text Assistance}

\begin{figure} 
\centering
\includegraphics[width=2.5 in]{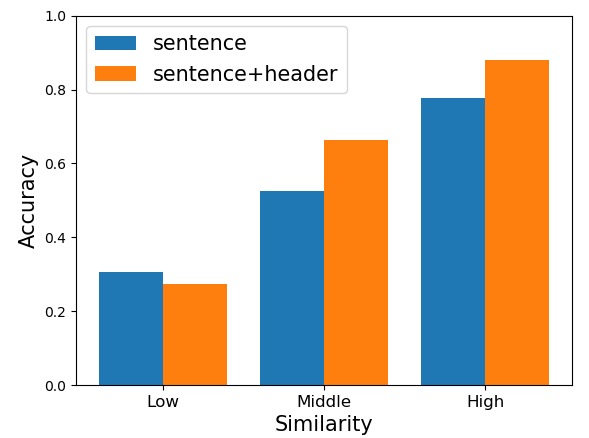}
\caption{Accuracy of the text assistance method.
}
\label{fig:similarity}
\end{figure}

To test the effectiveness of the text assistance method,
we randomly sample 4,000 pairs of insights and their most similar sentences in the reports, and ask 10 human annotators who are familiar with financial reports to label whether the pairs are of the same meaning.
The evaluation data is equally split into three groups according to their similarity scores.

As shown in figure~\ref{fig:similarity}, for both similarity functions, the higher similarity is the higher accuracy of their having the same meaning is.
Therefore we can use the similarity score as the ground truth of the insight importance.

In addition, we find that $\mathrm{Sim}_{s+h}$ performs better than $\mathrm{Sim}_{s}$.
It obtains nearly $90\%$ accuracy for the high similarity group.
The reason is that
$\mathrm{Sim}_{s+h}$ emphasizes the headers explicitly compared to the $\mathrm{Sim}_{s}$.
For the low similarity group, the accuracy of the similarity function with headers is lower than that with sentences, which indicates that the similarity function with headers is more distinguishable.

According to the human evaluation, in the later experiments we will use $\mathrm{Sim}_{s+h}$ as the similarity function.

\section{Experiment}

\subsection{Experiment Settings}

Based on the performance on the validation set, we set the embedding size to $64$ for the baseline methods and the proposed model. The vocabulary sizes in the financial report dataset and the SBNation dataset are 8,409 and 900, respectively.

The parameters are updated by Adam algorithm~\cite{DBLP:journals/corr/KingmaB14} on a single 1080 Ti GPU and initialized by sampling from the uniform distribution ($[-0.1, 0.1]$). The initial learning rate is $0.0003$. 
The model is trained in mini-batches with a batch size of $1$.

 
\subsection{Evaluation Metrics}

We report the ranking accuracy in three evaluation metrics: 
Mean Average Precision(mAP)@k, Normalized Discounted Cumulative Gain(NDCG)@k, and Precision@k. 
\begin{table*}
\caption{Evaluation results on financial report dataset. }
\centering
\begin{tabular}{l|ccc|cc|cc}
\hline
                & Precision@1 & Precision@3 & Precision@5 & mAP@3 & mAP@5 & NDCG@3 & NDCG@5 \\
\hline                
$Sig_{table}$   & 0.098 & 0.246 & 0.399 & 0.474 & 0.624 & 0.646 & 0.688 \\
$Sig_{dataset}$ & 0.107 & 0.249 & 0.408 & 0.473 & 0.621 & 0.649 & 0.692 \\
$Sig_{cluster}$ & \textbf{0.110} & \textbf{0.261} & \textbf{0.416} & \textbf{0.481} & \textbf{0.632} & \textbf{0.658} & \textbf{0.703} \\
\hline
$TAR_{cnn}$     & 0.118 & 0.278 & 0.444 & 0.525 & 0.686 & 0.738 & 0.757 \\
$TAR_{semantics}$  & 0.162 & 0.411 & 0.605 & 0.668 & 0.756 & 0.799 & 0.815 \\
$TAR_{memory}$  & \textbf{0.170} & \textbf{0.425} & \textbf{0.626} & \textbf{0.684} & \textbf{0.772} & \textbf{0.812} & \textbf{0.829} \\
\hline
\end{tabular}

\label{tab:fin}
\end{table*}


\subsection{Comparing Methods}
\label{comparing_methods}


We first compare three significant score calculation methods. The detailed calculation methods follow the definition of point insight and shape insight in \cite{tang2017extracting}.
\begin{itemize}
	\item{\boldmath $Sig_{table}$ calculates the significance from the data distributions in one table. It represents the insight importance when the insights are compared to other insights in the same table.}
    \item{\boldmath $Sig_{dataset}$ calculates the significance from the data distributions in all tables. We assume that all tables are inherently related to each other.
    } 
    \item{\boldmath $Sig_{cluster}$ first clusters the subspaces of all the insights in the dataset using the word embedding of the headers, then calculate the significance score of the data distributions in one cluster. We employ the K-Means method for clustering, and $k$ is set to $7$ for the best performance. 
    }
\end{itemize}

We also implement the Text Assisted Ranking (\textbf{TAR}) model with different components.
\begin{itemize}
    \item \boldmath $TAR_{cnn}$ adds the CNN to capture more statistical features in addition to the basic insight significance and insight type features. 
    \item \boldmath $TAR_{semantics}$ adds the table header as semantics information to the input in addition to the \boldmath $TAR_{cnn}$. 
    \item \boldmath $TAR_{memory}$ adds the memory component to the \boldmath $TAR_{semantics}$ to introduce the table context and relations among the insights. 
\end{itemize}



\subsection{Experiment Results and Analysis}
\label{experiment results}


\subsubsection{Financial Report Dataset}

Evaluation results on financial report dataset are shown in Table~\ref{tab:fin}.
In general, our proposed method achieves the best overall performance, which demonstrates its ability to fully explore the insight characteristics and modeling the insight importance.

We first compare the three baseline methods, $Sig_{table}$, $Sig_{dataset}$ and $Sig_{cluster}$, which calculate the significance scores in different ways. 
The performance of $Sig_{dataset}$ is slightly better than that of $Sig_{table}$, as the former method calculate the significance with a much larger space of data points.
The comparison result also supports our assumption that the statistical significance score method does not suit for small tables, as the significance score is unreliable while there are only very few insights from a table.
The cluster method $Sig_{cluster}$ achieves the best result, which demonstrates the importance of the header semantics since it is the clustering rule.
According to the result, we use the $Sig_{cluster}$ as significance score in the $TAR$ models.

A series of incremental experiments are conducted to evaluate the contributions of the key components in our proposed model. 
Three versions of $TAR$ model in incremental sequence, $TAR_{cnn}$, $TAR_{semantics}$ and $TAR_{memory}$, are provided. 
$TAR_{cnn}$ is a basic version that explores the insight type, the significance score and the subspace of insights. 
By introducing the subspace information, the $TAR_{cnn}$ model is exposed to more available information on the statistical data distribution instead of a single significance score, and slightly improves the ranking performance. 

Comparing to the gap between $TAR_{cnn}$ and $Sig_{cluster}$, the improvement between $TAR_{cnn}$ and $TAR_{semantics}$ is much more obvious. 
The result suggests that the semantics is an important factor when we determine the importance value of an insight.
Explicitly introducing the semantics largely enriches the insight representation space and improve the ranking performance significantly.

The $TAR_{memory}$ model, the complete version of our proposed model, achieves the best performance in all evaluation metrics. 
Compared with $TAR_{semantics}$, the $TAR_{memory}$ introduces the related insight information within one group for comparison.
The result supports our assumption that global table context and grouped insight relationship make a contribution to the process of insight ranking.

\subsubsection{Human Evaluation} 

\begin{table}
\caption{Human evaluation of top-k Precision.}
\small
\centering
\begin{tabular}{l|ccc}
\hline
  & Precision@1 & Precision@3 & Precision@5\\
\hline
$Sig_{cluster}$       & 0.727   & 0.629  & 0.540\\
$TAR_{memory}$        & \textbf{0.886}   & \textbf{0.813}   & \textbf{0.745}\textbf{}\\
\hline
\end{tabular}

\label{tab:human_accuracy}
\end{table}


We randomly sample 400 tables and ask the human experts to determine if the top-k insights and their most similar descriptions in the report are of the same meanings.
The result in Table~\ref{tab:human_accuracy}
implies that the recommendations of the insights according to our ranking model are of high accuracy and reliability.

\subsubsection{SBNation Dataset}

\begin{table}
\caption{Top-k Precision on SBNation dataset.}
\small
\centering
\begin{tabular}{l|cc}
\hline
  & Precision@1 & Precision@3 \\
\hline
$Sig_{cluster}$       & 0.503   & 0.513   \\
$TAR_{semantics}$     & 0.788   & 0.754   \\
$TAR_{memory}$        & \textbf{0.797}   & \textbf{0.759}   \\
\hline
\end{tabular}
\label{tab:nba}
\end{table}


The experimental result on SBNation dataset is shown in Table~\ref{tab:nba}. 
Different from the annual financial report dataset, the description in SBNation is much more rigid and lacks variation. 
Therefore we consider label matched sentences as the target, and mark insight importance as 0-or-1, either relevant or irrelevant, rather than continuous 0-to-1 values. 
NDCG and mAP cannot adapt to such labels in ranking problems. 
The value $k$ in Precision@k is set to 1 and 3, as the tables in SBNation are relatively smaller and most of them contain only 3 to 4 insights. 
Similar to the results in financial report dataset, the $TAR_{memory}$ achieves the best performance.

\subsection{Case Study}

\begin{table}
\caption{Case Study}
\small
\centering
\begin{tabular}{l|c|c}
\hline
Insight Descriptions & Gold & TAR \\
\hline
Collaboration and license revenue was 71.7  & 1 & 2 \\
million for the year ended, an increase of  &  &  \\
58.7 million compared to the year ended.    &   &   \\
\hline
General  and administrative expenses were  & 2 & 4 \\ 
27.8 million for the year ended, an increase&  &  \\
of 18.8 million compared to the year ended. &   &   \\                 
\hline
Research and development expenses were& 3 & 1 \\
58.6 million for the year ended, an increase&  &  \\
of 35.1 million compared to the year ended.    &   &   \\                 
\hline
We had 111 full-time employees including & 4 & 9 \\
82 employees engaged in development. &  &  \\       
\hline
The net valuation allowance increased by & 5 & 3 \\
4.9 million and 0.6 million respectively. &   &   \\              
\hline
\end{tabular}
\label{tab:case}
\end{table}


We present a ranking result example in Table~\ref{tab:case}. 
It consists of the top \textbf{5} insights in \textbf{10} insight candidates from one table.

The Precision@5 is $0.8$, a relatively high accuracy. 
The more detailed relative position of the top 5 insights is of less usefulness. 
Because the target ranking results only represent the probability of the insight's being in the text, and the importance of the top 5 insights are of little distinction.

The reason why the fourth insight is wrongly labeled is that the similarity score is incorrectly calculated and the gold standard is in fact inaccurate. 
We analyzed the insight description and found that the sentence is coincidently matched with a wrong insight because it contains some keywords in the headers and similar numbers. 
This serves as an example of the optimization direction of the text assisted approach. 
We would like to solve this problem by introducing the position of the sentence
to the text assistance to derive more accurate similarity function.

\section{Conclusion}

In this paper, we propose a context-aware memory network to rank the insight importance. 
The model explores the data characteristics and introduces table structure and semantics information into the ranking process. 
We construct a financial report dataset, in which the insight interestingness inferred from the human written description is used as annotated training data. 
Experimental results show that our approach largely improves the ranking precision.

In the future, we would like to 
investigate a more reliable similarity function to take the sentence position into account.
Also, instead of text assistance, we can explore more methods such as figure assistance and meta-data assistance to estimate the approximate score of the insight importance.

 

\bibliographystyle{aaai}
\bibliography{aaai19}

\end{document}